# Semi-Supervised Collective Classification
# via Hybrid Label Regularization

**Luke K. McDowell**                                                          LMCDOWEL@USNA.EDU

*Department of Computer Science, U.S. Naval Academy, Annapolis, MD U.S.A.*

**David W. Aha**                                                             DAVID.AHA@NRL.NAVY.MIL

*Navy Center for Applied Research in Artificial Intelligence, Naval Research Laboratory; Washington, DC U.S.A.*

## Abstract

Many classification problems involve data instances that are interlinked with each other, such as webpages connected by hyperlinks. Techniques for *collective classification* (CC) often increase accuracy for such data graphs, but usually require a fully-labeled training graph. In contrast, we examine how to improve the semi-supervised learning of CC models when given only a sparsely-labeled graph, a common situation. We first describe how to use novel combinations of classifiers to exploit the different characteristics of the relational features vs. the non-relational features. We also extend the ideas of *label regularization* to such hybrid classifiers, enabling them to leverage the unlabeled data to bias the learning process. We find that these techniques, which are efficient and easy to implement, significantly increase accuracy on three real datasets. In addition, our results explain conflicting findings from prior related studies.

## 1. Introduction

Collective classification (CC) often substantially increases classification accuracy when the class labels of inter-related objects are correlated (Jensen et al., 2004; Sen et al., 2008). Most work with CC performs learning using a fully-labeled training graph. However, acquiring such labels can be very difficult, and learning a classifier with only a few such labels can lead to very poor performance (Shi et al., 2011).

In response, a few researchers have recently examined the CC task where a classifier must be learned from a partially-labeled training graph, using some form of semi-supervised learning (SSL) to leverage the unlabeled portion of the graph. However, they have reported inconsistent or weak results, even when using the same datasets and similar algorithms. This includes Bilgic et al. (2010), who found moderate gains from SSL, whereas Shi et al. (2011) reported otherwise.

In this paper, we examine how to improve SSL learning for within-network CC when the provided graph is only sparsely labeled. We focus on traditional CC algorithms that learn a relational model of the data and then apply a collective inference algorithm such as the Iterative Classification Algorithm (ICA) or Gibbs sampling (Sen et al., 2008). Given the substantial number of recently proposed CC algorithms, we do not attempt here to establish the "best" CC algorithm for sparsely-labeled data. Rather, we ask: given a sparsely-labeled graph, *can some form of SSL significantly improve the accuracy of traditional CC?* If not (as argued by Shi et al. 2011), then these approaches may need to be entirely replaced with alternatives such as latent feature models (Tang & Liu, 2009) or label propagation (Shi et al., 2011). However, if SSL *can* be effective in this domain, then many challenges remain, but a substantial amount of existing research on CC can continue to be used and adapted.

In our studies, we confirm that the simplest forms of SSL do not perform consistently well for sparsely-labeled CC. However, we introduce two new techniques that are simple and computationally efficient, yet can significantly increase accuracy.

Our contributions are as follows. First, we show how the most relevant prior work can all be generalized into a single parameterized algorithm for semi-supervised CC, facilitating comparison. Second, we explain how to transform the node classifier used by an algorithm like ICA or Gibbs sampling into a "hybrid" classifier that uses one classifier for the non-relational features (i.e., the attributes of each node) and a different classifier for the relational features (i.e., those that depend on the links in the graph). This change enables novel combinations of classifiers with better performance. Third, we extend the idea of *label regularization* (Mann & McCallum, 2007) to support such hybrid classifiers. This technique uses predictions over the unlabeled data to induce models that better account for class distribution priors. Fourth, we demonstrate, using three standard datasets, that combining a hybrid classifier with label regularization leads to significant accuracy gains compared to existing SSL methods and other baselines. Finally, we use our results to explain the conflicting conclusions from previous studies.

*Table 1.* Related work on CC that has used some variant of semi-supervised ICA.

| | Node classifier | Hard/Soft | Learning alg. per Figure 1 | Data used with SSL-ICA |
|---|---|---|---|---|
| Shi et al. (2011) | Log. regression (LR) | Hard | ALL-ONEPASS | Citeseer |
| Bilgic et al. (2010) | Log. regression (LR) | Hard | KNOWN-ONEPASS | Cora, Citeseer |
| Lu & Getoor (2003) | Log. regression (LR) | Hard | ALL-EM | Cora, Citeseer |
| Xiang & Neville (2008) | RPT | Soft | KNOWN-EM | Gene, synthetic |

## 2. Background and Related Work

Assume we are given a graph $G = (V, E, X_A, Y, C)$ where $V$ is a set of nodes, $E$ is a set of edges, each $x_A \in X_A$ is an attribute vector for a node $v_i \in V$, each $Y_i \in Y$ is a label variable for $v_i$, and $C$ is the set of possible labels. We are also given a set of "known" values $Y^K$ for nodes $V^K \subset V$, so that $Y^K = \{y_i | v_i \in V^K\}$. Then the *within-network classification task* is to infer $Y^U$, the values of $Y_i$ for the remaining nodes $V^U$ with "unknown" values ($V^U = V \setminus V^K$).

For example, consider predicting whether a web page belongs to a professor or a student. Conventional approaches ignore the link relations and classify each page using the attributes $x_A$ derived from its content (e.g., words in the page). In contrast, methods for *collective classification* (Jensen et al., 2004) explicitly use the link structure by constructing additional relational features $x_R$ based on the labels of neighboring pages. For instance, one relational feature might count the number of pages labeled `Student` that are linked to each page. However, using such features is challenging, because some of the labels are initially unknown, and thus typically are estimated and then iteratively refined in some way. This can be done using algorithms such as belief propagation, Gibbs sampling, relaxation labeling, or ICA (Sen et al., 2008).

We focus on ICA, one of the simplest and most popular CC algorithms. ICA first predicts a label for every node in $V^U$ (the "unknown" nodes) using only the attributes $X_A$. Next, ICA constructs additional relational features $X_R$ using the known and predicted node labels ($Y^K$ and $Y^U$), then re-predicts labels for $V^U$ using both $X_A$ and $X_R$. This process of computing feature values and re-predicting labels is then repeated until convergence or for a fixed number of iterations.

### 2.1. Semi-supervised Collective Classification

Many CC variants, including ICA, use a classifier that predicts a node's label based on its attributes and (via relational features) the labels of linked nodes. In particular, ICA uses one "bootstrap classifier" that uses only the attributes ($M_A$) and one "node classifier" that uses both attributes and relational features ($M_{AR}$).

Most CC approaches assume that these classifiers are learned from a separate, fully-labeled training set. For our within-network task, however, we assume that there is a single sparsely-labeled graph. In this case, learning the classifiers $M_A$ and $M_{AR}$ is challenging because of label sparsity. Learning $M_{AR}$ is especially problematic, since relational features can only be used for learning in the rare case where *both* node endpoints of a link have known labels. For instance, if 10% of nodes are labeled, perhaps only 1% of links will qualify.

Given a large set of nodes but only a small set of provided labels, it is natural to consider some sort of semi-supervised learning (SSL). A few researchers have investigated this scenario, and Table 1 summarizes the most relevant studies, which all use some variant of semi-supervised ICA. We observe that all of these SSL variants can be generalized into a single SSL algorithm (see Figure 1), which we explain below.

These variants first learn an attribute-only classifier $M_A$ from the known labels, then predict labels for the unknown nodes $V^U$ with $M_A$ (steps 1-2). The known and predicted labels are then used to compute relational feature values (step 4). The variants then differ in how they use these values and in how many steps each variant takes. The simplest approach, taken by Shi et al. (2011), is to learn the classifier $M_{AR}$ using *all* of the labels, attributes, and relational feature values (step 5). Step 6 then uses $M_A$ and $M_{AR}$ to predict new

*Figure 1.* Generic SSL learning for CC. Variables with superscripts of "$K$" relate to "known nodes"; those with superscripts of "$U$" relate to nodes with unknown labels.

---

**SSL_learn** $(V, E, X, Y^K, n, LearnFromAll)=$

1  $M_A = learnClassifier(X_A^K, Y^K)$
2  $Y^U = predict(M_A, X_A^U)$
3  **for** $i = 1$ **to** $n$ **do**
4    $X_R = computeRelatFeatures(V, E, Y^K \cup Y^U)$
5    **if**$(LearnFromAll)$ // "All" variants
        $M_{AR} = learnClassifier(X_A, X_R, Y^K \cup Y^U)$
     **else**              // "Known" variants
        $M_{AR} = learnClassifier(X_A^K, X_R^K, Y^K)$
6    $Y^U = predict(M_A, M_{AR}, V, E, X_A^U, X_R^U, Y^K)$
7  **return** $Y^U$

---

labels (by executing ICA), and step 7 returns the set of predicted labels ($n = 1$). Bilgic et al. (2010) use the same approach, except that during step 5, they perform learning using *only* the known nodes $V^K$. This approach is still semi-supervised because the relational feature values for $V^K$ (as computed in step 4) are influenced by the predicted labels for the unknown nodes $V^U$. We call this latter variant KNOWN-ONEPASS, since it uses only the known nodes' labels for the actual learning and does one step of relational learning, and naturally call the former variant ALL-ONEPASS.

Alternatively, more complex variants perform a form of EM where the algorithm repeatedly estimates new labels given the current models (i.e., step 6 is the E-step) and then maximizes the probability of a new model given the current label estimates (i.e., steps 4-5 are the M-step). Based on the learning choice in step 5, this yields the variants ALL-EM and KNOWN-EM, which are approximately the approaches of Lu & Getoor (2003) and Xiang & Neville (2008), respectively. Lu & Getoor repeat based on a convergence condition, while Xiang & Neville use a fixed number of iterations (e.g., $n = 10$). Xiang & Neville also perform inference in step 6 (and learning) with a "soft" variant of ICA that uses probability estimates of the predicted labels, rather than choosing the most likely label for each node in $V^U$ (the "hard"-labeling approach used by the other variants). We use hard labeling; future work should compare these alternatives.

This prior work leaves three key problems unaddressed. First, there are conflicting results for whether semi-supervised ICA improves accuracy, with reports of no improvement (Shi et al., 2011), moderate improvement (Bilgic et al., 2010), or mixed results (Xiang & Neville, 2008). Lu & Getoor (2003) report substantial gains, but only consider graphs with at least 20% of the nodes labeled. Second, there is almost no comparison between the four SSL variants (or even discus-sion of the choices involved). One exception is Bilgic (2010), which finds KNOWN-ONEPASS to be superior to ALL-ONEPASS on two datasets, but does not investigate why. In contrast, we have generalized these algorithms in Figure 1, and study their relative performance in Section 5. Such comparisons aid understanding of the choices involved, and also establish the best baseline for future studies.

Finally, we find that none of the variants perform consistently well. Sections 3 and 4 describe the two key steps that we propose for improving their performance.

## 2.2. Other Approaches to Semi-supervised CC

Two additional relevant studies are Taskar et al. (2001) and Chu et al. (2006). However, these methods cannot handle cyclic graphs or have time complexity at least quadratic in the number of nodes ($N$), and thus do not scale to large, realistic graphs. The methods we use are only linear in $N$ (assuming realistic link densities), even with our improvements in Sections 3 and 4.

Others have explored how to perform within-network CC without needing to learn an explicit model for link-based features (the challenge for ICA discussed in Section 2.1). For instance, Tang & Liu (2009) use the links to create latent features that enable node classification without collective inference. Shi et al. (2011) propose label propagation based on derived latent links.

A few authors have proposed "relational-only" methods that perform no learning but use some label propagation or random walk to classify the nodes (e.g., Macskassy & Provost 2007). These variants can increase accuracy, but only for graphs that match their assumptions and have enough known labels.

As noted in Section 1, we do not attempt to compare against all of these methods but do use that of Macskassy & Provost, a common CC baseline.

## 3. Hybrid Classifiers for CC

Prior work with CC using ICA or Gibbs sampling has usually used a single node classifier (often logistic regression or Naive Bayes) to predict a label $y$ for each node based on the node's attributes ($x_A$) and relational features ($x_R$). Instead, we propose to use two distinct classifiers that make separate predictions based on $x_A$ and $x_R$. If we assume that $x_A$ and $x_R$ are conditionally independent given the class label $y$, we can then compute the combined prediction

$$p(y|x) = p(y|x_A, x_R) = \frac{p(x_A|y)p(x_R|y)p(y)}{p(x_A, x_R)}$$

$$= \frac{\frac{p(y|x_A)p(x_A)}{p(y)} \frac{p(y|x_R)p(x_R)}{p(y)} p(y)}{p(x_A, x_R)}$$

$$= \alpha \frac{p(y|x_A)p(y|x_R)}{p(y)} \tag{1}$$

where $\alpha$ is a normalizing constant independent of $y$.

Using such a "hybrid" classifier has two main advantages. First, this method allows us to choose different types of classifiers for the attributes vs. the relational features. For instance, most prior work (e.g., Sen et al. 2008) has found that logistic regression (LR) performs best overall for CC. However, McDowell et al. (2009) found that "multiset" relational features usually performed best, but are incompatible with the vector-based representation of LR. Combining LR with attributes plus Naive Bayes (NB) with multiset relational features yields a new $LR+NB$ classifier that resolves this representational conflict and may increase accuracy. Second, hybrid classifiers may increase accuracy, even when the two classifiers are the same type (e.g., $LR+LR$, as we show and explain in Section 5.

Combining two different classifiers to make a single prediction is a special case of an ensemble. A few papers have considered how to apply some type of ensemble for CC. For instance, Preisach & Schmidt-Thieme (2008) use an ensemble to combine predictions based on different link types, while Eldardiry & Neville (2011) use an ensemble after each step of collective inference. These studies combine multiple classifiers using voting, stacking, or averaging, rather than via a probabilistic rule like Equation 1. Also, they do not use the unlabeled data for SSL; instead, they use fully-labeled training data or relational-only algorithms.

The work most closely related to ours is Lu & Getoor (2003), which uses an ensemble of two LR classifiers, combined similarly to Equation 1. They informally state that this approach outperformed a single LR classifier with SSL, but did not explain why or report comparisons. In contrast, our paper is the first to specifically demonstrate that a hybrid LR+LR classifier can often improve accuracy dramatically, and Section 5 explains why. In addition, this paper is the first to propose the $LR+NB$ combination, and we show that it can perform particularly well for some datasets.

## 4. Adding Label Regularization

Label regularization (Mann & McCallum, 2007) is designed to make SSL more robust by encouraging a learned LR classifier (or other exponential family model) to produce probability estimates on the unlabeled data so that the resultant class distribution resembles an expected distribution. More specifically, let $\tilde{p}(y)$ be the expected label distribution, which can be computed from the training data. Let $\hat{p_\theta}(y)$ be the empirical distribution, which is computed over the unlabeled part of the training data, given the current parameter settings $\theta$, as follows:

$$\hat{p_\theta}(y) = \frac{1}{|X^U|} \sum_{x \in X^U} p_\theta(y|x).$$

Mann & McCallum then augment the traditional LR objective function with an additional term $\lambda\Delta(\tilde{p}, \hat{p_\theta})$ based on the KL-divergence between $\tilde{p}$ and $\hat{p_\theta}$:

$$\Delta(\tilde{p}, \hat{p_\theta}) = \sum_{y_j \in Y} \tilde{p}(y_j) \log \frac{\tilde{p}(y_j)}{\hat{p_\theta}(y_j)}.$$

They (and we) set the tuning parameter $\lambda = 10 \times |V^K|$. The new term $\lambda\Delta(\tilde{p}, \hat{p_\theta})$ penalizes parameter settings where there is a large difference between the expected distribution and empirical distribution. Thus, it uses the unlabeled data to help choose more plausible parameter values and avoid degenerate cases (e.g., where most nodes are assigned to the same class label).

Label regularization could be directly applied to CC classifiers based on non-hybrid LR (though to our knowledge this has not been done previously). However, this would not exploit the advantages we later demonstrate for hybrid classifiers. Given a hybrid LR+LR classifier, we could apply label regularization separately to each classifier. However, this would not ensure that the combined predictions given by Equation 1 resemble the desired label distribution, nor would this work for hybrid combinations not based on LR, such as $LR+NB$. Instead, we use the following strategy to adapt label regularization to hybrid classifiers. First, we learn $p(y|x_R)$ (the relational classifier) ignoring the attributes and label regularization. Then, we treat $p(y|x_R)$ as fixed and define $\beta_y = \alpha \frac{p(y|x_R)}{p(y)}$, which allows us to simplify Equation 1 to

$$p(y|x) = p(y|x_A, x_R) = \beta_y p(y|x_A).$$

Next, we assume that the attribute-based classifier is a standard multinomial logistic regression model, so

$$p_\theta(y|x_A) = \frac{exp(x_A^T \theta_y)}{\sum_{y' \in C} exp(x_A^T \theta_{y'})}$$

where $\theta_y$ is a parameter vector for class $y$. Combining the previous two equations and normalizing yields

$$p_\theta(y|x) = \frac{1}{Z} \beta_y exp(x_A^T \theta_y)$$

where, since $\sum_y p_\theta(y|x) = 1$, $Z = \sum_{y'} \beta_{y'} exp(x_A^T \theta_{y'})$.

Gradient methods can now be used to optimize the log-likelihood of the training data, with the term $\Delta(\tilde{p}, \hat{p}_\theta)$ in the objective providing label regularization. Thus, we now compute the gradient of this term. For each node $v_i$, let $x_A$ be the values of $v_i$'s attributes, and $x_k$ be the $k^{th}$ such attribute value. Then the gradient with respect to $\theta_{y,k}$ (the parameter associated with the $k^{th}$ attribute for class $y$) is

$$\frac{\partial \Delta}{\partial \theta_{y,k}} = \frac{\partial}{\partial \theta_{y,k}} \sum_{y' \in C} \tilde{p}(y') \log \frac{\tilde{p}(y')}{\hat{p}_\theta(y')}$$

$$= -\sum_{y' \in C} \frac{\tilde{p}(y')}{\hat{p}_\theta(y')} \frac{\partial}{\partial \theta_{y,k}} \hat{p}_\theta(y')$$

$$= \frac{-1}{|X^U|} \sum_{x \in X^U} \sum_{y' \in C} \frac{\tilde{p}(y')}{\hat{p}_\theta(y')} \frac{\partial}{\partial \theta_{y,k}} p_\theta(y'|x)$$

$$= \frac{-1}{|X^U|} \sum_{x \in X^U} \left[ \frac{\tilde{p}(y')}{\hat{p}_\theta(y')} \frac{x_k (Z\beta_y e^{x_A^T \theta_y} - (\beta_y e^{x_A^T \theta_y})^2)}{Z^2} \right.$$

$$\left. + \sum_{y' \in C\backslash y} \frac{\tilde{p}(y')}{\hat{p}_\theta(y')} \frac{0 - (x_k \beta_y e^{x_A^T \theta_y}) \beta_{y'} e^{x_A^T \theta_{y'}}}{Z^2} \right]$$

$$= -\frac{1}{|X^U|} \sum_{x \in X^U} \left[ \frac{\tilde{p}(y)}{\hat{p}_\theta(y)} x_k p_\theta(y|x)(1 - p_\theta(y|x)) \right.$$

$$\left. + \sum_{y' \in C\backslash y} \frac{\tilde{p}(y')}{\hat{p}_\theta(y')} (-x_k) p_\theta(y|x) p_\theta(y'|x) \right]$$

$$= \sum_{x \in X^U} \frac{x_k p_\theta(y|x)}{|X^U|} \left[ \sum_{y' \in C} \frac{\tilde{p}(y')}{\hat{p}_\theta(y')} p_\theta(y'|x) - \frac{\tilde{p}(y)}{\hat{p}_\theta(y)} \right].$$

This approach can work for any hybrid classifier that includes LR in the combination, including *LR+LR* and *LR+NB*. In contrast, recent work (Mann & McCallum, 2010) extended label regularization to support conditional random fields, but did not consider the kind of hybrid classifiers that we examine here.

# 5. Experimental Study

## 5.1. Datasets and Features

Prior studies used at most two non-synthetic datasets with semi-supervised ICA; we used the union of their datasets (see Tables 1 & 2). We removed all nodes with no links, but we did not (like some others did) use only the largest connected component of the graphs.

**Cora** (see Sen et al. 2008) is a collection of machine learning papers. **Citeseer** (see Sen et al.) is a collection of research papers drawn from the Citeseer collection. For both datasets, the attributes represent the presence or absence of particular words, and citations provide links between the documents. We ignored link direction, as with Bilgic et al. (2010). They also report substantially higher accuracies using principal component analysis (PCA) to reduce the dimensionality of the attributes, so to provide a stronger baseline we mimic their setup and use the 100 top attribute features after applying PCA to the entire graph.

*Table 2.* Data sets summary.

| Characteristics | **Cora** | **CiteSeer** | **Gene** |
|---|---|---|---|
| Total nodes | 2708 | 3312 | 1243 |
| Total links | 5278 | 4536 | 1672 |
| Class labels | 7 | 6 | 2 |
| % dominant class | 16% | 21% | 56% |

**Gene** (see Jensen et al. 2004) describes the yeast genome at the protein level; links represent protein interactions. We mimic Xiang & Neville (2008) and predict protein localization using four attributes: Phenotype, Class, Essential, and Chromosome. We binarized these attributes, yielding 54 binary attributes.

For relational features, LR classifiers used "proportion" features, which compute the fraction of a node's neighbors that have label $y$, as done by Bilgic et al. (2010). NB classifiers instead used "multiset" features which record the distribution of labels in each node's neighborhood, then use conditional independence assumptions to update the estimated probabilities based on each such label. Previous work found such features to perform best for NB (McDowell et al., 2009).

## 5.2. Node Classifier and Regularization

Prior work has usually found LR to be superior to NB for CC (Sen et al., 2008; Bilgic et al., 2010) and therefore we always use LR for (at least) the attribute-based classification. For the node classifier $M_{AR}$, we evaluate five options: *LR* is a single (non-hybrid) classifier that uses logistic regression. *LR+LR* is a hybrid classifier that uses two LR classifiers, while *LR+LR+Reg* adds label regularization. *LR+NB* is a hybrid classifier that uses LR for the attributes and NB for the relational features, and *LR+NB+Reg* adds label regularization.

For standard regularization (separate from label regularization), we used a Gaussian prior with LR parameters, with variance $\sigma^2$ chosen as described below. For NB, we used a Dirichlet prior on each feature with $\alpha$ chosen as described below (McDowell et al., 2009).

Of the four studies listed in Table 1, none specify how regularization parameters (if any) were chosen. For sparsely-labeled data, we observed that these choices can have a large impact on accuracy. To ensure fair comparisons, we used five-fold cross-validation on the labeled data (learning also had access to the unlabeled data), selecting the value that maximized accuracy on the held-out labeled data. For *LR+LR* classifiers, we normalized all features, which allowed us to use a single value of $\sigma^2$ for both classifiers. For *LR+NB*, we first found $\sigma^2$ for LR, then estimated $\alpha$ for NB.

### 5.3. Learning Algorithms

We evaluate four variants of semi-supervised ICA (see Section 2): ALL-EM, ALL-ONEPASS, KNOWN-EM, and KNOWN-ONEPASS. For each, step 6 of the algorithm executes ICA with 10 iterations, and the EM-variants use $n = 10$ iterations of the main loop.

We compare against three baselines. The first, NO-SSL, is like KNOWN-ONEPASS in applying ICA one time using both attributes and relational features, but NO-SSL learns the node classifier without using any unlabeled data (i.e., with no SSL). The second, ATTR-ONLY, predicts the unknown labels only once, using an attribute-only classifier that ignores unlabeled data while learning. The third, RELAT-ONLY, is a standard relational-only baseline (the wvRN+RL classifier of Macskassy & Provost 2007) that repeatedly estimates labels based on the labels of all linked neighbors. These three baselines never use label regularization.

### 5.4. Evaluation Procedure

We report accuracy averaged over 15 trials. For each trial, we randomly selected some fraction of the nodes (the "label density") to be "known" nodes $V^K$. The remaining nodes $V^U$ have unknown labels and form the test set part of graph $G$. We focus on the sparsely-labeled case where the density is less than 10%.

To assess significance, we use paired t-tests with a 5% significance level. However, the test sets are not disjoint across the 15 trials, and thus a traditional paired t-test may yield false conclusions. To compensate for this effect, we use the recently described methodology of Wang et al. (2011), which was shown to reduce false positives to the expected level. This makes our results more conservative compared to uncorrected t-tests.

### 5.5. Results

We first consider the best learning algorithm (ALL-EM, as shown later) with various classifiers, then compare different learning algorithms with the best node classifier (*LR+NB+Reg*). Finally, we use our findings to explain the results of previous studies.

**Result 1: Using a hybrid classifier and label regularization each increases accuracy, and combining the two techniques yields the best overall results.** Figure 2 shows the average accu-

racy, for ALL-EM, as label density is varied from 1% to 50%. Below, we discuss results only for the sparse case (density less than 10%). Each line represents a different node classifier, and symbols indicate (some of the) significant differences (see caption).

The hybrid *LR+LR* almost always outperforms *LR*, with especially large gains for Cora. Likewise, changing the classifier to *LR+NB* almost always yields some additional gain, even when the label density is as high as 9%. Furthermore, adding label regularization (with *LR+LR+Reg* or *LR+NB+Reg*) always outperforms *LR+LR*, and always matches or exceeds the accuracy of *LR+NB*. The gains from label regularization are sometimes substantial, especially when the label density is low, but remain large even at a density of 9% for Gene. For instance, for Citeseer when the density is 1%, label regularization improves *LR+LR* by 13.1% and *LR+NB* by 9.6% (both significantly).[1]

Overall, *LR+NB+Reg* performs best and its accuracy is never lower than the alternatives. It significantly outperforms *LR*, and often substantially outperforms the other classifiers, especially vs. those without label regularization and/or when the label density is lower.

Without label regularization, some learning runs converged on degenerate distributions, with one class disproportionately represented. Adding label regularization substantially reduced this problem, as intended. We also found that, compared to *LR*, *LR+LR* learned much more reasonable weights $\theta$ for the relational features (e.g., that match our knowledge of the actual link-based correlations). When there are few labeled nodes, the LR optimization routine may find a model where the attributes alone explain the data well, with small weights for the relational features. Placing these features in a separate model (as with *LR+LR*) ensures that they will be more heavily used, increasing accuracy since both attributes and relational features are informative for these datasets.

**Result 2: ALL-EM outperforms the baselines and usually the other SSL variants.** Table 3 compares the four SSL algorithms and the three baseline algorithms, using *LR+NB+Reg*. ALL-EM is the most consistent, and is always one of the best algorithms (with one exception for Gene). In many cases, ALL-EM's gains are significant vs. the other methods.

ALL-ONEPASS and KNOWN-ONEPASS outperform

---

[1]Other results (not shown) confirm that *LR+LR+Reg* (or *LR+NB+Reg*) also outperformed a (non-hybrid) *LR* with label regularization added. For example, for density 1%-9%, *LR+NB+Reg* had (usually significant) gains of 1.1-7.7% (Cora), 0.3-2.6% (Citeseer), and 3.4-12.3% (Gene).

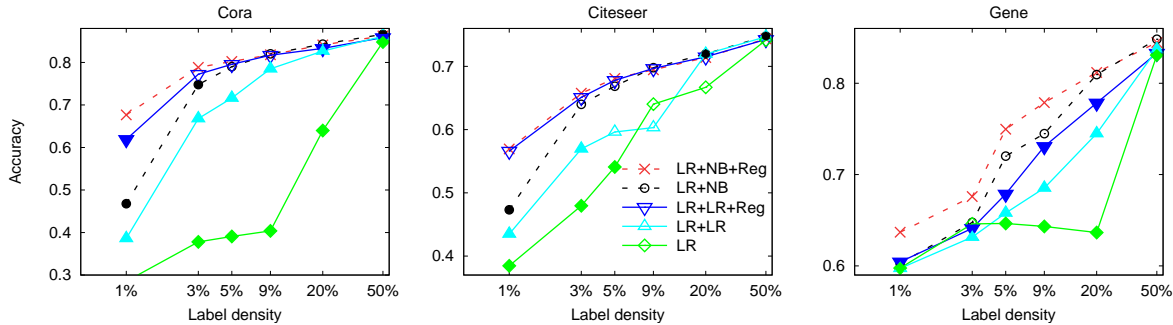

*Figure 2.* Average accuracy using the ALL-EM learning algorithm. Symbol shapes that are filled in (not hollow) indicate where that performance was significantly *worse* than *LR+NB+Reg*, based on a corrected, paired t-test (see Section 5.4). To facilitate comparison, variants based on *LR+NB* are shown with dashed lines. The x-axis uses a log scale.

NO-SSL by varying degrees, but KNOWN-EM is much worse, never matching the best accuracy and often under-performing NO-SSL. For this data, the repeated EM iterations seem to be a poor choice when the actual learning is performed over only the small number of "known" nodes; KNOWN-ONEPASS performs better.

**Result 3: Without hybrid classification and label regularization, accuracy decreases substantially.** Table 4 compares the same SSL algorithms shown in Table 3, but now using the simpler *LR* classifier. This setting is closest to that of Bilgic et al. (2010) and Shi et al. (2011). To save space, we show only Cora; trends are similar with Citeseer and Gene. Compared to Table 3's results with *LR+NB+Reg*, performance with *LR* usually drops substantially (as already partly seen in Figure 2). The ALL variants suffer the biggest drops, especially ALL-EM, because without a good hybrid classifier and label regularization, learning may be based on estimated labels with a degenerate distribution. The KNOWN variants are more insulated from this effect because poor label estimates only affect the relational feature values (not the labels directly seen by the learning algorithm).

**Discussion:** Table 4 showed that using *LR* changes the relative performance of the SSL algorithms compared to when using *LR+NB+Reg*. These results help us explain the previously discussed conflicting findings related to Table 1. First, with a simple *LR* classifier, KNOWN-ONEPASS outperforms ALL-ONEPASS (and provides reasonable gains over NO-SSL) – consistent with Bilgic et al. (2010). Second, ALL-ONEPASS behaves similarly to NO-SSL – the same poor behavior that led Shi et al. (2011) to reject semi-supervised ICA. Note that both of these conclusions change when a better classifier like *LR+NB+Reg* is used (see Table 3). Third, KNOWN-EM behaves reasonably well but not the best; this may explain why Xiang & Neville found gains for this algorithm only in some cases. Finally,

Lu & Getoor reported strong results with ALL-EM and a classifier like *LR+LR*, tested for label densities of at least 20%. For lower densities, our results (see Figure 2) show that this combination is not as strong.

Overall, our results show that, while the simplest forms of semi-supervised ICA do not perform well, using a hybrid classifier with label regularization enables the most sophisticated learning algorithm (ALL-EM) to work well, leading to significant accuracy gains. To quantify the overall impact of our changes, the bottom row of Table 3 shows results with a natural benchmark: the basic *LR* algorithm with KNOWN-ONEPASS. This SSL algorithm worked best with *LR* and mimics the setup of Bilgic et al. (2010), who reported spending considerable effort to improve their performance. Comparing vs. the top row of Table 3 (*LR+NB+Reg* with ALL-EM), we find consistent and mostly significant gains, ranging from 4.3-26.8% for Cora, 0.9-12.7% for Citeseer, and 2.7-3.4% (with one loss of -0.8%) for Gene. Thus, we find consistent gains for our new methods compared to this benchmark method.

## 6. Conclusion

We have generalized the algorithms of multiple previous studies of semi-supervised ICA, explained their performance trends, and demonstrated that a hybrid classifier with label regularization can significantly increase accuracy compared to alternative approaches. For our data, the *LR+NB* combination performed best, though this will not hold for every dataset. However, our hybrid approach enables any combination of probabilistic classifiers to be selected based on the data characteristics. Moreover, our extension of label regularization can also be used to increase accuracy, as long as one of the classifiers is in the exponential family (like LR). Such hybrid classifiers with label regularization may be useful in many other tasks, including across-network CC or non-relational classification.

*Table 3.* Average accuracy with $LR+NB+Reg$, where each row uses a different learning algorithm. (The last row is an exception; it uses $LR$ with Known-OnePass, mimicking Bilgic et al. 2010.) Within each column, the best value is in bold and a star indicates significantly *worse* accuracy vs. All-EM. Relat-Only is the wvRN+RL baseline; see Section 5.3.

| | Cora | | | | Citeseer | | | | Gene | | | |
|---|---|---|---|---|---|---|---|---|---|---|---|---|
| Label Density | 1% | 3% | 5% | 9% | 1% | 3% | 5% | 9% | 1% | 3% | 5% | 9% |
| All-EM | **67.7** | **78.9** | **80.2** | **81.8** | **56.9** | **65.8** | **68.1** | **69.5** | **63.7** | 67.6 | **75.0** | **77.9** |
| All-OnePass | 57.2* | 76.0* | 78.1* | 80.9* | 52.3* | 64.1* | 67.9 | **69.5** | 61.7* | 67.4 | 71.2* | 72.8* |
| Known-EM | 40.6* | 59.6* | 64.8* | 70.8* | 45.2* | 58.1* | 62.4* | 66.6* | 59.0* | 64.8 | 58.2* | 59.5* |
| Known-OnePass | 51.7* | 71.1* | 75.4* | 79.7* | 48.7* | 62.0* | 66.0* | 68.4* | 63.6 | **69.2** | 73.9 | 76.2 |
| No-SSL | 37.0* | 56.3* | 64.0* | 72.7* | 40.6* | 55.8* | 63.2* | 68.8 | 60.0* | 67.4 | 70.4* | 73.7* |
| Attr-Only | 37.7* | 54.7* | 60.0* | 65.1* | 41.0* | 55.8* | 62.3* | 66.4* | 60.4* | 64.8 | 68.6* | 70.2* |
| Relat-Only | 42.6* | 63.7* | 72.4* | 77.2* | 32.0* | 47.9* | 52.9* | 55.3* | 55.4* | 63.5 | 66.5* | 70.1* |
| Bilgic et al. baseline | 43.6* | 64.5* | 71.4* | 77.5* | 44.2* | 59.5* | 64.5* | 68.6 | 61.0 | 68.4 | 71.6* | 74.8* |

*Table 4.* Average accuracy for Cora with (non-hybrid) $LR$

| Label Density | 1% | 3% | 5% | 9% |
|---|---|---|---|---|
| All-EM | 28.5 | 37.8 | 39.1 | 40.4 |
| All-OnePass | 36.4 | 57.5 | 64.2 | 70.3 |
| Known-EM | 41.3 | 61.8 | 68.7 | 74.4 |
| Known-OnePass | **43.6** | **64.5** | **71.4** | **77.5** |
| No-SSL | 37.0 | 56.3 | 64.0 | 72.7 |

Our results need to be confirmed with additional datasets, and we will explore alternatives to ICA such as Gibbs sampling and soft-labeling methods, and the use of other hybrid combination rules such as stacking. Moreover, the best algorithms should be compared against other methods discussed in Section 2.2. However, our results already show that there is more potential for semi-supervised learning based on ICA than was suggested by earlier studies, and we have presented two techniques that improve its performance.

## Acknowledgments

Thanks to Alexandra Olteanu, Li Pu, Majid Yazdani, and the anonymous referees for comments that helped to improve this work. Portions of this analysis used Proximity, an open-source software environment from the Univ. of Massachusetts, Amherst. This work was supported in part by NSF award number 1116439.